\documentclass[runningheads]{llncs}

\usepackage{graphicx}
\usepackage{comment}
\usepackage{amsmath,amssymb} % define this before the line numbering.
\usepackage{color}
\usepackage[width=122mm,left=12mm,paperwidth=146mm,height=193mm,top=12mm,paperheight=217mm]{geometry}

\usepackage{microtype}
\usepackage{soul}
\usepackage{url}
\usepackage[hidelinks]{hyperref}
\usepackage[utf8]{inputenc}
\usepackage{graphicx}
\usepackage{amsmath}
\usepackage{booktabs}
\usepackage{algorithm}
\usepackage{amsfonts}
\usepackage{algorithm}
\usepackage{algpseudocode}
\usepackage{graphicx}
\usepackage{textcomp}
\usepackage{xspace}
\usepackage{subfig}

\newcommand{\commentout}[1]{}
            
\newcommand{\xiaowei}[1]{{\color{blue}#1}}

\newcommand{\hana}[1]{{\color{magenta}#1}}

\newcommand{\network}{\mathcal{N}}

\usepackage{color}

\newcommand{\tool}[1]{\textsc{#1}\xspace}
\newcommand{\deepcover}{\tool{DeepCover}}

\newcommand{\gradcam}{{\sc g}rad{\sc cam}\xspace}
\newcommand{\lime}{{\sc lime}\xspace}
\newcommand{\shap}{{\sc shap}\xspace}
\newcommand{\rise}{{\sc rise}\xspace}
\newcommand{\extremal}{{\sc e}xtremal\xspace}

\begin{document}

\pagestyle{headings}
\mainmatter
\def\ECCVSubNumber{6192}  % Insert your submission number here
\title{Explaining Image Classifiers\\ using Statistical Fault Localization}

\author{Youcheng~Sun\orcidID{0000-0002-1893-6259}\inst{1} \and
        Hana~Chockler\orcidID{0000-0003-1219-0713}\inst{2} \and
        Xiaowei~Huang\orcidID{0000-0001-6267-0366}\inst{3} \and
        Daniel~Kroening\orcidID{0000-0002-6681-5283}\inst{4}
}

\institute{Queen's University Belfast \and King's College London \and University of Liverpool \and University of Oxford}

\maketitle

\begin{abstract}
The black-box nature of deep neural networks (DNNs) makes it impossible to understand \emph{why}
a particular output is produced, creating demand for ``Explainable AI''.
In~this paper, we show that statistical fault localization (SFL) techniques from software engineering
deliver high quality explanations of the outputs of DNNs, where we define an explanation as a minimal
subset of features sufficient for making the same decision
as for the original input. We present an algorithm and a tool called 
\deepcover, which synthesizes a ranking of the features of the inputs using
SFL and constructs explanations for the decisions of the DNN based on this ranking. 
We compare explanations produced by \deepcover with those of the state-of-the-art tools
\gradcam, \lime, \shap, \rise and \extremal and show that explanations generated by \deepcover are 
consistently better across a broad set of experiments.
On a benchmark set with known ground truth, 
\deepcover achieves $76.7\%$ accuracy, which is $6\%$ better than the second best \extremal.

\end{abstract}

\section{Introduction}
\label{sec:introduction}

Deep neural networks (DNNs) are increasingly used in place of traditionally engineered software in many areas.
DNNs are complex non-linear functions with algorithmically generated (and not engineered) coefficients, and
therefore are effectively ``black boxes''. They are
given an input and produce an output, but the calculation of these outputs is difficult to explain~\cite{rahwan2019machine}.
The goal of \textit{explainable AI} is to create artifacts that provide a rationale for why a neural network generates a particular output for a particular input. This is argued to enable stakeholders to understand and appropriately trust neural networks. %~\cite{eu-ai-guidance}.

A typical use-case of DNNs is classification of highly dimensional inputs, such as images. DNNs are multi-layered networks with a predefined structure that consists of layers of neurons. The coefficients for the neurons are determined by a training process
on a data set with given classification labels. %~\cite{Goodfellow2016}.
The standard criterion for the adequacy of training is the accuracy of the network on a separate validation data set. %~\cite{assess a DNN - cite a standard process}.
This criterion is clearly only as comprehensive as the validation data set. In~particular,
this approach suffers from the risk that the validation data set is lacking an important instance~\cite{ziegler2016google}. %,lambert2016understanding}.
Explanations provide additional insight into the decision process of a neural
network~\cite{DARPA,olah2018the}. 

%Explanations can be used to guide the training process to the missing inputs and
%to signal when the decisions are sufficiently accurate.

%Explanations have been claimed to address this problem by providing additional insight into the decision process of a neural
%network~\cite{gunning2017explainable,olah2018the}. Explanations can be used to guide the training process to the missing inputs and to signal when the decisions are sufficiently accurate. 

\commentout{
\paragraph*{State of the art} The state of the art in producing explanations for image classifiers is an approach called  \textit{SHapley Additive exPlanations (SHAP)}~\cite{lundberg2017unified}, which assigns an ``importance value'' to each pixel.
The algorithm treats the classification of a multi-dimensional input as a multi-player collaborative game, where each player represents a dimension. The importance value of a pixel is the contribution it makes to the classification. This method provides a reasonable, and accurate, explanation based on game theory. In practice, owing to the computational complexity of the algorithm, the implementation approximates the solution, leading to inaccuracies. In addition, since SHAP combines multiple techniques that are conceptually different, including~\cite{lime,deeplift,datta2016algorithmic}, it is intrinsically complex.
}

In traditional software development, SFL measures have a substantial track record of helping engineers to
debug sequential programs~\cite{naish2011model}. %lucia2014extended} %,naish2011model}. %, lucia2014extended,landsberg2015evaluation}. 
These measures rank program locations by
counting the number of times a particular location is visited in passing and in failing executions for a
given test suite and applying statistical formulae. The ranked list is presented to the engineer.
The main advantage of SFL measures is that they are comparatively inexpensive to compute.
%The accuracy of their ranking clearly depends on the quality of the test suite as well as the statistical formula.
%~\cite{} %wong2010family, zhang2017theoretical, perez2017test, jiang2011practical, santelices2009lightweight, feldt2016test}.
There are more than a hundred of measures in the literature~\cite{wong2016survey}.
Some of the most widely used measures are Zoltar, Ochiai, Tarantula and Wong-II~\cite{zoltar,ochiai1957zoogeographic,jones2005empirical,wong2007effective}.

\subsubsection{Our contribution}  

We propose to apply the concept of \emph{explanations} introduced by Halpern and Pearl in the context of
\emph{actual causality}~\cite{HP05}. 
%definition of actual causality~\cite{HP05,HP05a}. 
Specifically, we define an explanation as a subset of features of the input that is \emph{sufficient}
(in terms of explaining the cause of the outcome),
\emph{minimal} (i.e., not containing irrelevant or redundant elements), and \emph{not obvious}. 

Using this definition and SFL measures, we have developed {\bf \deepcover} -- a tool that provides explanations for DNNs that classify images. \deepcover ranks the pixels using four well-known SFL measures (Zoltar, Ochiai, Tarantula and Wong-II) based on the results of running test suites constructed from random mutations of the input image. \deepcover then uses this ranking to efficiently construct an approximation of the explanation (as explained below, the exact computation is intractable).

We compare the quality of the explanations produced by \deepcover with
those generated by the state-of-the-art tools \gradcam, \lime, \shap, \rise and
\extremal in several complementary scenarios. 
First, we measure the size of the explanations as an indication of the quality
of the explanations. To complement this setup, we further apply the explanation tools
to the problem of weakly supervised object localization (WSOL). We also create a
``chimera'' benchmark, consisting of images with a known ground truth.
\deepcover exhibits consistently better performance in these evaluations.
Finally, we investigate the use of explanations in a DNN security application, and show that
\deepcover successfully identifies the backdoors that trigger Trojaning attacks.

\section{Related Work}

There is a large number of methods for explaining DNN decisions.
Our approach belongs to a category of methods that compute local perturbations.
Such methods compute and visualize the important features of an input instance to
explain the corresponding output. Given a particular input, \lime~\cite{lime} samples the the
neighborhood of this input and creates a linear model to approximate the model's local behavior;
owing to the high computational cost of this approach, the ranking uses super-pixels instead of
individual pixels. In~\cite{datta2016algorithmic}, the natural distribution of the input is replaced by a user-defined distribution and the Shapley Value method is used to analyze combinations of input features and to rank their importance.  In~\cite{chen2018learning}, the importance of input features is estimated by measuring the the flow of information between inputs and outputs. Both the Shapley Value and the information-theoretic approaches are computationally expensive.  In RISE~\cite{petsiuk2018rise}, the importance of a pixel is computed as the expectation over all local perturbations conditioned on the event that the pixel is observed. More recently, the concept of ``extreme perturbations'' has been introduced to improve the perturbation analysis by the \extremal algorithm~\cite{fong2019understanding}.

%Another thread of research back-propagates a machine model's output signal (layer by layer) towards the input and estimates the associated importance of each input feature. 
On the other hand, gradient-based methods only need one backward pass. \gradcam~\cite{CAM} passes the class-specific gradient into the final convolutional layer of a DNN to coarsely highlight important regions of an input image. %In \cite{wagner2019interpretable}, the descent direction is computed via integrated gradients instead of the normal gradient. 
In~\cite{shrikumar2017learning}, the activation of each neuron is compared with some reference point, and its contribution score for the final output is assigned according to the difference. The work of~\cite{lime,datta2016algorithmic,shrikumar2017learning,lundberg2017unified} is similar:
an approximation of the model's local behavior using a simpler linear model and an application of the Shapley Value theory to solve this model. 

Our algorithm for generating explanations is inspired by the  statistical fault localization (SFL) techniques in software testing~\cite{naish2011model} %lucia2014extended,naish2011model}
(see Sec.~\ref{sec:SFL} for an overview).
SFL measures have the advantage of being simple and efficient. They are widely used for localizing causes of software failures.
%Different from correlation coefficient approach for measuring the statistical relationship, SFL does not assume a probability distribution, but rather a test suite, which does not represent any probability distribution. In fact, it purposefully does not represent the probability distribution, because, in particular, SFL benefits from a balanced test suite even if errors happen very rarely on random inputs. 
Moreover, there are single-bug optimal measures~\cite{landsberg2015evaluation} that guarantee that the fault is localized when it is the single cause for the program failure. While it is not always possible to localize a single best feature to explain a DNN image classifier, single-bug optimal measures often perform well even when there is more than one fault in the program~\cite{landsberg2018optimising}.
From the software engineering perspective, our work can be regarded as applying SFL techniques for diagnosing the neural network's
decision. This complements recent works on the testing and validation of
AI~\cite{sun-concolic,odena2019tensorfuzz,sun2020reliability,noller2020hydiff},
for which a detailed survey can be found in \cite{huang2020survey}.

%Explanations may also be useful to assess security aspects of DNNs. \cite{trojaning} propose a Trojan attack on DNNs by embedding a trigger into the input images that activates the malicious behavior of the model. Our experiments show that our tool localizes the Trojan trigger with high accuracy.
%In this paper, we regard the Trojan trigger as the explanation for the neural network's malicious behavior and we apply our tool to localize it.

\section{Preliminaries}
\label{sec:preliminaries}
\subsection{Deep neural networks (DNNs)}
\label{sec:dnns}

We briefly review the relevant definitions of deep neural networks. Let $f: \mathcal{I} \rightarrow \mathcal{O}$ be a deep neural network $\network$ with $N$~layers. For a given input $x\in \mathcal{I}$, 
$f(x)\in\mathcal{O}$ calculates the output of the DNN, which could be, for instance, a classification label. Images are among the most popular inputs for DNNs, and in this paper we focus on DNNs that classify images.  Specifically, we have
\begin{equation}\label{eq:dnn}
f(x) = f_N(\ldots f_2(f_1(x;W_1,b_1);W_2,b_2)\ldots ;W_N,b_N)
\end{equation}
where $W_i$ and $b_i$ for $i = 1,2,\ldots,N$ are learnable parameters, and $f_i(z_{i-1};W_{i-1},\\b_{i-1})$
is the layer function that maps the output of layer $(i-1)$, i.e., $z_{i-1}$, to the input of layer~$i$. 
The combination of the layer functions yields a highly complex behavior, and the analysis of the
information flow within a DNN is challenging. There is a variety of layer functions for DNNs,
including fully connected layers, convolutional layers and max-pooling layers.
Our algorithm is independent of the specific internals of the DNN and treats a given DNN as a black box.

\subsection{Statistical fault localization (SFL)}
\label{sec:SFL}

Statistical fault localization techniques (SFL)~\cite{naish2011model}, %lucia2014extended}, %\cite{naish2011model, lo2010comprehensive, lucia2014extended, zou2019empirical, wong2010family, zhang2017theoretical, perez2017test, jiang2011practical, santelices2009lightweight, feldt2016test, wong2016survey, zoltar, ochiai1957zoogeographic, jones2005empirical, wong2007effective, landsberg2015evaluation}
have been widely used in software testing to aid in locating the
causes of failures of programs. SFL techniques rank program elements (e.g., statements or assignments)
based on their \emph{suspiciousness scores}. Intuitively, a program element is more suspicious if it appears in failed
executions more frequently than in correct executions (the exact formulas for ranking differ). 
Diagnosis of the faulty program can then be conducted by manually examining the ranked list of elements
in descending order of their suspiciousness
until the culprit for the fault is found. 

The SFL procedure first executes the program under test using a set of inputs. It records the program executions
together with a set of Boolean flags that indicate whether a particular element was executed by the current test.
%as \emph{program spectra}, meaning that the execution is instrumented to modify a set of Boolean flags that indicate whether a particular program %element was executed. 
The task of a fault localization tool is to compute a ranking of the program elements based on the values of these flags.
%program spectra. 
Following the notation in~\cite{naish2011model}, the suspiciousness score of each program statement $s$ is calculated
from a set of parameters $\langle a^s_\mathit{ep}, a^s_\mathit{ef}, a^s_\mathit{np}, a^s_\mathit{nf} \rangle$ that
give the number of times the statement $s$ is executed ($e$) or not executed ($n$) on passing ($p$)
and on failing ($f$) tests. For instance, $a^s_\mathit{ep}$ is the number of tests that passed and executed $s$.

A large number of measures have been proposed to calculate the suspiciousness scores.
In Eq.~\ref{ex:measures} we list the most widely 
used ones~\cite{ochiai1957zoogeographic,zoltar,jones2005empirical,wong2007effective};
those are also the measures that we use in our ranking procedure. 

\vspace{-2ex}
\noindent\begin{subequations}\label{ex:measures}%
\begin{tabular}{@{\hspace{-0.2cm}}l@{}l}%
\begin{minipage}{6.6cm}%
\begin{equation}\text{Ochiai:}\,\,\, \frac{a_\mathit{ef}^s}{\sqrt{(a_\mathit{ef}^s+a_\mathit{nf}^s)(a_\mathit{ef}^s+a_\mathit{ep}^s)}}\label{eq:ochiai}\end{equation}
\end{minipage}
& \begin{minipage}{5.8cm}%
\begin{equation}\text{Tarantula:}\,\,\,  \displaystyle\frac{\frac{a_\mathit{ef}^s}{a_\mathit{ef}^s+a_\mathit{nf}^s}}{\frac{a_\mathit{ef}^s}{a_\mathit{ef}^s+a_\mathit{nf}^s}+\frac{a_\mathit{ep}^s}{a_\mathit{ep}^s+a_\mathit{np}^s}}\label{eq:tarantula}\end{equation}
\end{minipage}\\
\begin{minipage}{6.6cm}%
\begin{equation}\text{Zoltar:}\,\,\, \frac{a_\mathit{ef}^s}{a_\mathit{ef}^s+a_\mathit{nf}^s+a_\mathit{ep}^s+\frac{10000a_\mathit{nf}^sa_\mathit{ep}^s}{a_\mathit{ef}^{s}}}\end{equation}
\end{minipage}
& \begin{minipage}{5.8cm}%
\begin{equation}
\text{Wong-II:} \label{eq:wong-ii}\,\,\,   a_\mathit{ef}^s-a_\mathit{ep}^s
\end{equation}
\end{minipage}
\end{tabular}
\end{subequations}
\vspace{1ex}

%\begin{subequations}\label{ex:measures}
%\begin{align}
%&\text{Ochiai\cite{ochiai1957zoogeographic}:}\,\,\, \frac{a_\mathit{ef}^s}{\sqrt{(a_\mathit{ef}^s+a_\mathit{nf}^s)(a_\mathit{ef}^s+a_\mathit{ep}^s)}}\label{eq:ochiai}\\
%&\text{Zoltar\cite{zoltar}:}\,\,\, \frac{a_\mathit{ef}^s}{a_\mathit{ef}^s+a_\mathit{nf}^s+a_\mathit{ep}^s+\frac{10000a_\mathit{ef}^sa_\mathit{ep}^s}{a_\mathit{ef}^{s}}}\\
%&\text{Tarantula\cite{jones2005empirical}:}\,\,\,  \displaystyle\frac{\frac{a_\mathit{ef}^s}{a_\mathit{ef}^s+a_\mathit{nf}^s}}{\frac{a_\mathit{ef}^s}{a_\mathit{ef}^s+a_\mathit{nf}^s}+\frac{a_\mathit{ep}^s}{a_\mathit{ep}^s+a_\mathit{np}^s}}\label{eq:tarantula}\\
%&\text{Wong-II\cite{wong2007effective}:} \label{eq:wong-ii}\,\,\,   a_\mathit{ef}^s-a_\mathit{ep}^s
%\end{align}
%\end{subequations}
%

\noindent
There is no single best measure for fault localization. Different measures perform better on different applications, and best practice is to use them together.

\section{What is an Explanation?}\label{sec:explain}

An explanation of an output of an automated procedure is essential in many areas, including verification,
planning, diagnosis and the like. A good explanation can increase a user's confidence
in the result. Explanations are also useful for determining whether there is a fault in the automated procedure:
if the explanation does not make sense, it may indicate that the procedure is faulty. 
It is less clear how to define what a \emph{good} explanation is. There have been a number of definitions of explanations over the years in various domains of computer science~\cite{CH97,Gar88,Pea88}, philosophy~\cite{Hem65} and statistics~\cite{Sal89}.
The recent increase in the number of machine learning applications and the advances in deep learning led to the need
for \textit{explainable AI}, which is advocated, among others, by DARPA~\cite{DARPA} to promote understanding, trust, and adoption of future autonomous systems based on learning algorithms (and, in particular, image classification DNNs).
DARPA provides a list of questions that a good explanation should answer and an epistemic state of the user after receiving a good explanation. The description of this epistemic state boils down to \emph{adding useful information} about the output of the algorithm
and \emph{increasing trust} of the user in the algorithm.
 
In this paper, we are loosely adopting the definition of explanations by Halpern and Pearl~\cite{HP05}, which is based on their definition of actual causality~\cite{HP05a}. Roughly speaking, they state that a good explanation gives an answer to the question
\textit{``why did this outcome occur''}, which is similar in spirit to DARPA's informal description.
As we do not define our setting in terms of actual causality, we omit the parts of the definition that refer to
causal models and causal settings. The remaining parts of the definition of explanation are:
\begin{enumerate}
    \item an explanation is a \textit{sufficient} cause of the outcome;
    \item an explanation is a \textit{minimal} such cause (that is, it does not contain irrelevant or redundant elements);
    \item an explanation is \textit{not obvious}; in other words, before being given the explanation,
    the user could conceivably imagine other explanations for the outcome.
\end{enumerate}

In image classification using DNNs, the non-obviousness holds for all but extremely trivial images.
Translating $1)$ and $2)$ into our setting, we get the following definition.
\begin{definition}\label{def:explanation}
An explanation in image classification is a minimal subset of pixels of a given input image that is sufficient for the DNN to
classify the image, where ``sufficient'' is defined as containing only this subset of pixels from the original image,
with the other pixels set to the background colour.
\end{definition}
We note that (1) the explanation cannot be too small (or empty), as a too small subset of pixels would violate the sufficiency requirement,
and (2) there can be multiple explanations for a given input image.

The precise computation of an explanation in our setting is intractable, as it is equivalent to the earlier definition of
explanations in binary causal models, which is DP-complete~\cite{EL04}.
%It is easy to see that the precise computation of an explanation according to Def.~\ref{def:explanation} is
%intractable, as the problem
%is NP-complete
%(the proof of equivalence is deferred to the full paper).
A brute-force approach of checking all subsets of
pixels of the input image is exponential in the size of the image. 
In~Sec.~\ref{sec:proposed} we describe an efficient linear-time approach to computing an approximation of an explanation and argue that
for practical purposes, this approximation is sufficiently close to an exact explanation as defined above.

\section{SFL Explanation for DNNs}
\label{sec:proposed}

We propose a \textit{black-box explanation technique} based on statistical fault localization. 
In~traditional software development, SFL measures are used for ranking program elements that cause a failure.
In our setup, the goal is different: we are searching for an explanation of why a particular input to a given
DNN yields a particular output; our technique is agnostic to whether the output is correct.
We start with describing our algorithm on a high level and then present the pseudo-code and technical details.

\paragraph{Generating the test suite}
SFL requires test inputs. Given an input image $x$ that is classified by the DNN $\mathcal{N}$ as $y = \mathcal{N}[x]$, we generate a set of images by \textit{randomly mutating}~$x$.
A \emph{legal mutation} masks a subset of the pixels of~$x$, i.e., sets these pixels to the background color.
The DNN computes an output for each mutant; we annotate it with ``$y$'' if that output matches that of $x$,
and with ``$\neg{y}$'' to indicate that the output differs.
The resulting test suite $T(x)$ of annotated mutants is an input to the \deepcover algorithm. 

\paragraph{Ranking the pixels of $x$}
We assume that the original input $x$ consists of $n$ pixels
$\mathcal{P}=\{p_1,\dots,p_n\}$. Each test input $t\in T(x)$ exhibits a particular spectrum
for the pixel set, in which some pixels are the same as in the original input $x$ and others are masked.
The presence or masking of a pixel in $x$ may affect the output of the DNN.

We use SFL measures to rank the set of pixels of $x$ by slightly abusing the notions of passing and failing tests.
For a pixel $p_i$ of $x$, we compute the vector $\langle a^i_\mathit{ep}, a^i_\mathit{ef}, a^i_\mathit{np}, a^i_\mathit{nf} \rangle$ as follows:
\begin{itemize}
    \item $a^i_\mathit{ep}$ is the number of mutants in $T(x)$ labeled $y$ in which $p_i$ is not masked;
    \item $a^i_\mathit{ef}$ is the number of mutants in $T(x)$ labeled $\neg{y}$ in which $p_i$ is not masked;
    \item $a^i_\mathit{np}$ is the number of mutants in $T(x)$ labeled $y$ in which $p_i$ is masked;
    \item $a^i_\mathit{nf}$ is the number of mutants in $T(x)$ labeled $\neg{y}$ in which $p_i$ is masked.
\end{itemize}

Once we construct the vector $\langle a^i_\mathit{ep}, a^i_\mathit{ef}, a^i_\mathit{np}, a^i_\mathit{nf} \rangle$ for every pixel,
we apply the SFL measures discussed in Sec.~\ref{sec:SFL} to rank the pixels of $x$ for their
importance regarding the DNN's output (the importance corresponds to the suspiciousness score computed by SFL measures).

\paragraph{Constructing an explanation}
An explanation is constructed by iteratively adding
pixels to the set in the descending order of their ranking (that is, we start with the highest-ranked pixels)
until the set becomes sufficient for the DNN to classify the image. This set 
is presented to the user as an explanation.

\subsection{SFL explanation algorithm}
\label{sec:sfl-explanation}
\commentout{
The computation of an SFL explanation for a given DNN
is described in Algorithm~\ref{algo:sbe}. Given the DNN $\mathcal{N}$,
a particular input $x$ and a particular fault localization measure $M$,
it synthesizes the subset of pixels $\mathcal{P}^\mathit{exp}$ that is presented to the user as an explanation.
}
We now present our algorithms in detail. Algorithm~\ref{algo:sbe} starts by 
calling procedure\\
$\mathit{test\_inputs\_gen}$ to generate the
set $T(x)$ of test inputs (Line $1$). It~then computes the vector
$\langle a_\mathit{ep}^i, a_\mathit{ef}^i, a_\mathit{np}^i, a_\mathit{nf}^i \rangle$ for each pixel $p_i\in\mathcal{P}$
using $T(x)$ (Lines $2$--$5$). Next, the algorithm computes the ranking of each pixel
according to the specified measure $M$ (Line $6$).
Formulas for measures are as in Eq.~\eqref{eq:ochiai}--\eqref{eq:wong-ii}.
The pixels are sorted in descending order of their ranking (from high
$\mathit{value}$ to low $\mathit{value}$). 

\begin{algorithm}[t]
  \caption{SFL Explanation for DNNs}
  \label{algo:sbe}
  \begin{flushleft}
    \textbf{INPUT:} DNN $\mathcal{N}$, image $x$, SFL measure $M$\\
    \textbf{OUTPUT:} a subset of pixels $\mathcal{P}^\mathit{exp}$
  \end{flushleft}
  \begin{algorithmic}[1]
  %\State $s\leftarrow$ number of test inputs
   \State $T(x)\leftarrow \mathit{test\_inputs\_gen}(\mathcal{N},x)$
   \For{each pixel $p_i \in \mathcal{P}$}
      \State calculate $a_\mathit{ep}^i, a_\mathit{ef}^i, a_\mathit{np}^i, a_\mathit{nf}^i$ from $T(x)$
      \State $\mathit{value}_i\leftarrow M(a_\mathit{ep}^i, a_\mathit{ef}^i, a_\mathit{np}^i, a_\mathit{nf}^i)$
   \EndFor
   \State $\mathit{pixel\_ranking} \leftarrow$ pixels in $\mathcal{P}$ from high $\mathit{value}$ to low
   \State $\mathcal{P}^\mathit{exp}\leftarrow\emptyset$
   \For{each pixel $p_i \in \mathit{pixel\_ranking}$}
        \State $\mathcal{P}^\mathit{exp}\leftarrow\mathcal{P}^\mathit{exp}\cup\{p_i\}$
        \State $x^\mathit{exp}\leftarrow$ mask pixels of $x$ that are \textbf{not} in $\mathcal{P}^\mathit{exp}$
        \If{$\mathcal{N}[x^\mathit{exp}]=\mathcal{N}[x]$} 
          \State \Return{$\mathcal{P}^\mathit{exp}$}
        \EndIf
   \EndFor
  \end{algorithmic}
\end{algorithm}

From Line~$7$ onward in Algorithm~\ref{algo:sbe}, we construct a subset of pixels
$\mathcal{P}^\mathit{exp}$ to explain $\mathcal{N}$'s output on this particular input $x$ as follows.
We add pixels to $\mathcal{P}^\mathit{exp}$, while $\mathcal{N}$'s output on $\mathcal{P}^\mathit{exp}$ does not
match $\mathcal{N}[x]$. This process terminates when $\mathcal{N}$'s output is the same
as on the whole image $x$. 
%The explanation subset of pixels $\mathcal{P}^{exp}$ grows, from high valued
%pixels towards low valued pixels, until a proper explanation satisfying the Definition \ref{def:explanation}
%is found.
Finally, $\mathcal{P}^\mathit{exp}$ is returned as the explanation.
At the end of this section we discuss why $\mathcal{P}^\mathit{exp}$ is not
a precise explanation according to Def.~\ref{def:explanation} and argue that it is a good approximation
(coinciding with a precise explanation in most cases).

As the quality of the ranked list computed by SFL measures inherently depends on the quality of the test suite,
the choice of the set $T(x)$ of mutant images plays an important role in our SFL
explanation algorithm for DNNs. While it is beyond the scope of this paper to identify the best set
$T(x)$, we propose an effective method for generating $T(x)$ in Algorithm~\ref{algo:tx_gen}. 
The core idea of Algorithm~\ref{algo:tx_gen} is to balance the number of test inputs annotated with
``$y$'' (that play the role of the passing traces) with the number of test inputs annotated with ``$\neg{y}$''
(that play the role of the failing traces). Its motivation is that, when applying fault localisation in software
debugging, the rule of thumb is to maintain a balance between passing and failing cases.

The fraction $\sigma$ of the set of pixels of $x$ that are going to be masked in a mutant is
initialized by a random or selected number between $0$ and $1$ (Line~$2$) and is later updated at each iteration according to the
decision of $\mathcal{N}$ on the previously constructed mutant. 
In each iteration of the algorithm, a~randomly chosen set of ($\sigma\cdot n$) pixels in $x$
is masked and the resulting new input $x'$ is added to $T(x)$ (Lines~$4$--$5$). 
Roughly speaking, if a mutant is not classified with the same label as $x$, we decrease the fraction of masked pixels by a pre-defined small number $\epsilon$; if the mutant is classified with the same label as $x$, we increase the fraction of masked pixels by the same~$\epsilon$.
%If the newly
%generated $x'$ has a different DNN output (line 6), then at next iteration, we tend to find
%another test input that is likely to reserve $y$. This is why at line 7, the value of $\sigma$,
%which controls the portion of pixels to be masked, is decreased by some pre-defined threshold
%$\epsilon$. Alternatively, when $\mathcal{N}[x']=\mathcal{N}[x]$, we tend to find another that differs
%the DNN output from $y=\mathcal{N}[x]$, by changing more pixels (line 9).

\begin{algorithm}[t]
  \caption{$\mathit{test\_inputs\_gen}(\mathcal{N}, x)$}
  \label{algo:tx_gen}
  \begin{flushleft}
    \textbf{INPUT:} DNN $\mathcal{N}$, image $x$  (with $n$ pixels)\\
    \textbf{OUTPUT:} test suite $T(x)$\\
    \textbf{PARAMETERS:} $\sigma$, $\epsilon$, test suite size $m$\\
  \end{flushleft}
    \begin{algorithmic}[1]
    
    \State $T(x)\leftarrow\emptyset$
    \State $\sigma\leftarrow$ sample in the range $(0,1)$
    \While{$|T(x)|<m$}
      \State $x'\leftarrow$ randomly select and mask $\sigma\cdot n$ pixels in $x$
      \State $T(x)\leftarrow T(x)\cup \{x'\}$
      \If{$\mathcal{N}[x']\neq \mathcal{N}[x]$}
      \State $\sigma\leftarrow\max\{\sigma-\epsilon,0\}$
      \Else 
      \State $\sigma\leftarrow\min\{\sigma+\epsilon,1\}$
      \EndIf
    \EndWhile
    
    \State \Return{$T(x)$}
    
    \end{algorithmic}
\end{algorithm}

\subsection{Relationship between $\mathcal{P}^\mathit{exp}$ and Def.~\ref{def:explanation}}
Recall that Def.~\ref{def:explanation} requires an explanation to be \emph{sufficient}, \emph{minimal}, and \emph{not obvious}
(see Sec.~\ref{sec:explain}).
As we argued above, the non-obviousness requirement holds for all but very simple images. 
It is also easy to see that $\mathcal{P}^\mathit{exp}$ is sufficient, 
since this is a stopping condition for adding pixels to this set (Line $11$ in Algorithm~\ref{algo:sbe}).

The only condition that might not hold is minimality.
The reason for possible non-minimality is that
the pixels of $x$ are added to the explanation in the order of their ranking, with the highest-ranking pixels being added first.
It~is therefore possible that there is a high-ranked pixel that was added in one of
the previous iterations, but is now not necessary for the correct classification of the image (note that the process of adding
pixels to the explanation stops when the DNN successfully classifies the image; this, however, shows minimality only with respect to
the order of addition of pixels). We believe that the redundancy resulting from our approach is likely to be small,
as higher-ranked pixels have a larger effect on the DNN's decision.
In fact, even if our explanation is, strictly speaking, not minimal, it might not be a disadvantage,
as it was found that humans prefer explanations with some redundancy~\cite{zemla2017evaluating}. 

Another advantage of our algorithm is that its running time is linear in the size of the set $T(x)$ and the size of the image,
hence it is much more efficient than the brute-force computation of all explanations as described in Sec.~\ref{sec:explain} (and in fact, any algorithm
that computes a precise explanation, as the problem is intractable).
One hypothetical advantage of the enumeration algorithm is that it can produce all explanations; however, multiple explanations do not necessarily provide better insight into the decision process.

\section{Experimental Evaluation}\label{sec:evaluation}

We have implemented the SFL explanation algorithm for DNNs presented in Sec.~\ref{sec:proposed} 
in the tool \deepcover\footnote{https://github.com/theyoucheng/deepcover}.
We now present the experimental results.
We tested \deepcover on a variety of DNN models for ImageNet and 
we compare \deepcover with the most popular and most recent work 
in AI explanation:  \lime~\cite{lime}, 
\shap~\cite{lundberg2017unified}, \gradcam~\cite{CAM},  \rise~\cite{petsiuk2018rise} and \extremal~\cite{fong2019understanding}.\footnote{\lime version 0.1.33; \shap version 0.29.1; \gradcam, \rise and \extremal are from
https://github.com/facebookresearch/TorchRay (commit 6a198ee61d229360a3def590410378d2ed6f1f06)}

\commentout{
\begin{itemize}
    \item We tested \deepcover on a variety of DNN models for ImageNet, including Xception, VGG16, MobileNet and VGG Face.
    \item We compare \deepcover with the most popular and most recent work in AI explanation:  \lime~\cite{lime}, \gradcam~\cite{CAM}, \shap~\cite{lundberg2017unified}, \rise~\cite{petsiuk2018rise} and \extremal~\cite{fong2019understanding}.
    \item To address the groundtruth problem, we have synthesised images with proxied groundtruth.
    \item We demonstrate the power of \deepcover to detect DNN Trojaning attacks.
\end{itemize}
}
\subsection{Experimental setup}
We configure the heuristic test generation in Algorithm~\ref{algo:tx_gen} with $\sigma=\frac{1}{5}$ and $\epsilon=\frac{1}{6}$, and the size $m$ of the test set $T(x)$ is $2$$,$$000$.
These values have been chosen empirically and remain the same through all experiments. It is possible that
they are not appropriate for all input images, and that for some inputs increasing $m$ or tuning $\sigma$ and $\epsilon$ produces a better explanation. All experiments are run on a laptop with a 3.9\,GHz Intel i7-7820HQ and 16\,GB of memory.

\subsection{Are the explanations from \deepcover useful?}

Fig. \ref{fig:xception} showcases representative output from \deepcover on the Xception model.
We can say that explanations are indeed useful and meaningful. Each subfigure in Fig. \ref{fig:xception}
provides the original input and the output of \deepcover. We highlight misclassifications and counter-intuitive explanations
in red.
One of the more interesting examples is the ``cowboy hat''image. Although Xception labels the input image correctly,
an explanation produced by \deepcover indicates that this decision is not based on the correct feature (the hat in the image),
but on the face, which is an unexpected feature for the label `cowboy hat'. While this image was not, technically speaking,
misclassified, the explanation points to a flaw in the DNN's reasoning.
The ``wool'' and ``whistle'' are two misclassifications by Xception, and the explanations generated by \deepcover
can help us to understand why the misclassification happens: there are similarities between the features that are used for
the correct and the incorrect labels.

\newcommand{\lab}[1]{\scriptsize\sffamily{}#1}

\begin{figure}
\centering
\begin{minipage}{.55\textwidth}
  \centering
  \subfloat[{\lab{\color{red}`cowboy hat'}}]{
    \includegraphics[width=0.1\linewidth]{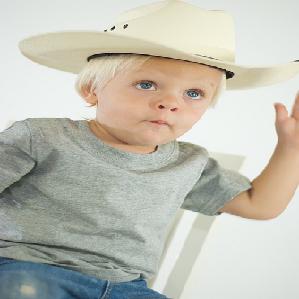}\hfill
    \includegraphics[width=0.1\linewidth]{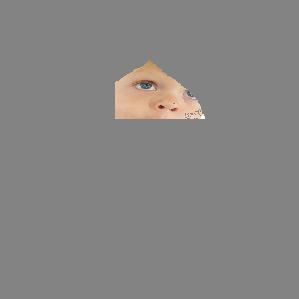}
  }\hspace{0.cm}
  \subfloat[\lab{`dog'}]{
    \includegraphics[width=0.1\linewidth]{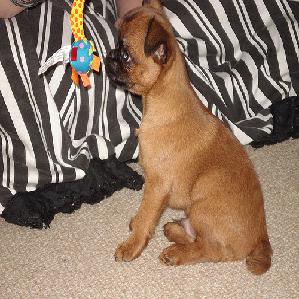}\hfill
    \includegraphics[width=0.1\linewidth]{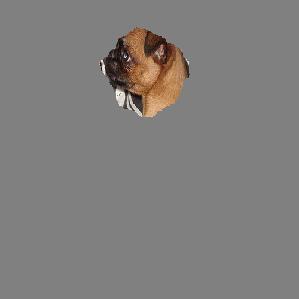}
  }\hspace{0.cm}
  \subfloat[{\lab`numbfish'}]{
    \includegraphics[width=0.1\linewidth]{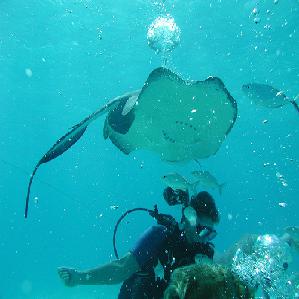}\hfill
    \includegraphics[width=0.1\linewidth]{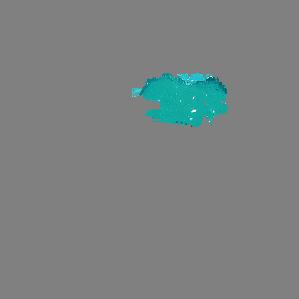}
  }\hspace{0.cm}
  \subfloat[{\lab`sheep'}]{
    \includegraphics[width=0.1\linewidth]{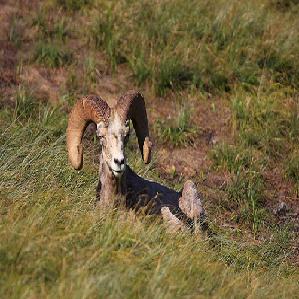}\hfill
    \includegraphics[width=0.1\linewidth]{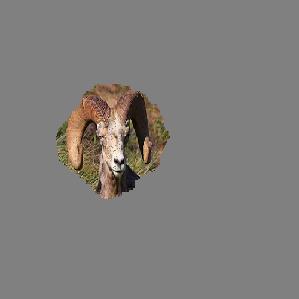}
  }\hspace{0.cm}\\
  \subfloat[\lab{`hare'}]{
    \includegraphics[width=0.1\linewidth]{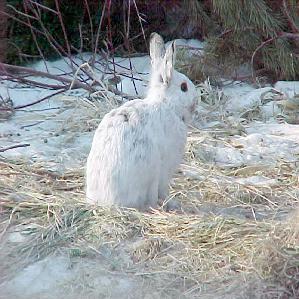}\hfill
    \includegraphics[width=0.1\linewidth]{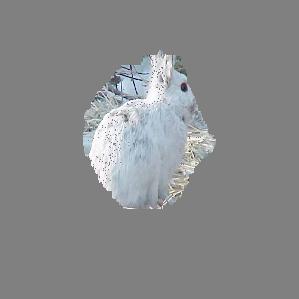}
  }\hspace{0.cm}
  \subfloat[\lab{`mushroom'}]{
    \includegraphics[width=0.1\linewidth]{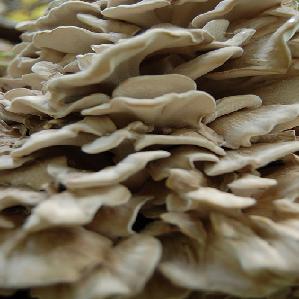}\hfill
    \includegraphics[width=0.1\linewidth]{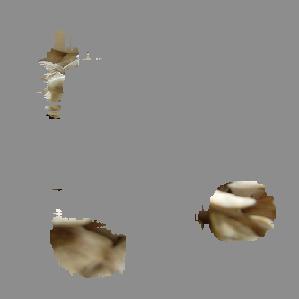}
  }\hspace{0.cm}
  \subfloat[\lab{\color{red}`wool'}]{
    \includegraphics[width=0.1\linewidth]{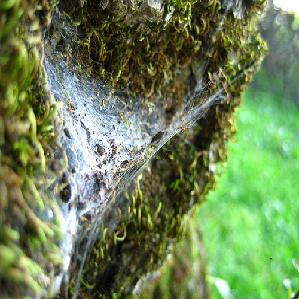}\hfill
    \includegraphics[width=0.1\linewidth]{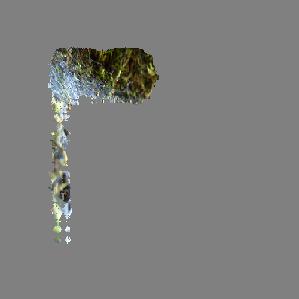}
  }\hspace{0.cm}
  \subfloat[\lab{`turnstile'}]{
    \includegraphics[width=0.1\linewidth]{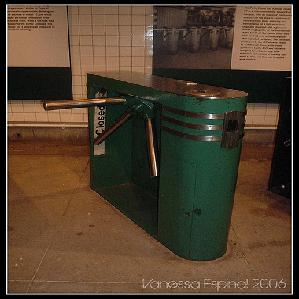}\hfill
    \includegraphics[width=0.1\linewidth]{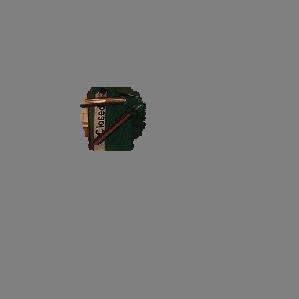}
  }\hspace{0.cm}\\
  \subfloat[\lab{`langur'}]{
    \includegraphics[width=0.1\linewidth]{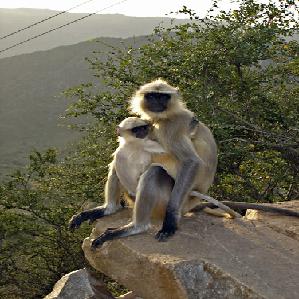}\hfill
    \includegraphics[width=0.1\linewidth]{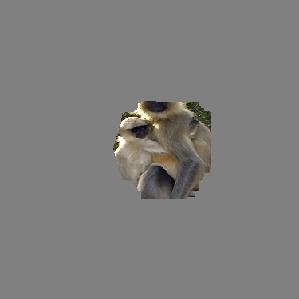}
  }\hspace{0.cm}
  \subfloat[\lab{\color{red}`whistle'}]{
    \includegraphics[width=0.1\linewidth]{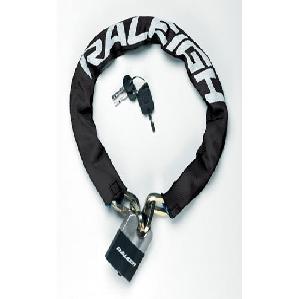}\hfill
    \includegraphics[width=0.1\linewidth]{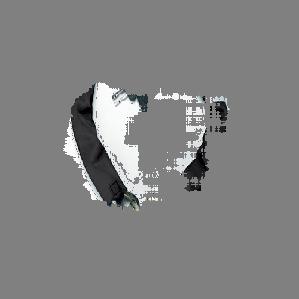}
  }\hspace{0.cm}
  \subfloat[\lab{`unicycle'}]{
    \includegraphics[width=0.1\linewidth]{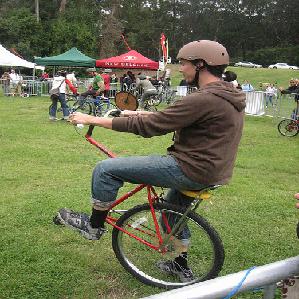}\hfill
    \includegraphics[width=0.1\linewidth]{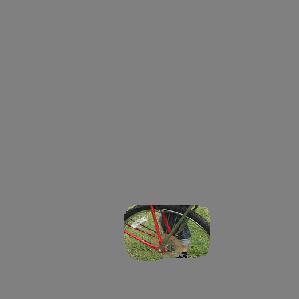}
  }\hspace{0.cm}
  \subfloat[\lab{`fire engine'}]{
    \includegraphics[width=0.1\linewidth]{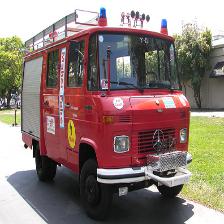}\hfill
    \includegraphics[width=0.1\linewidth]{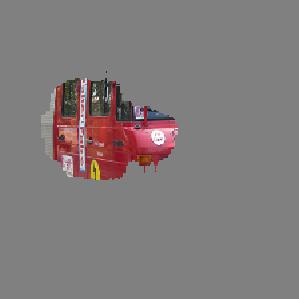}
  }\hspace{0.cm}\\
  \subfloat[\lab{`traffic light'}]{
    \includegraphics[width=0.1\linewidth]{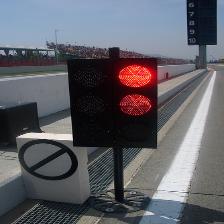}\hfill
    \includegraphics[width=0.1\linewidth]{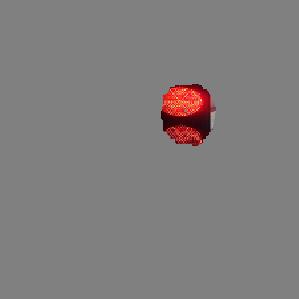}
  }\hspace{0.cm}
  \subfloat[\lab{`ballpoint'}]{
    \includegraphics[width=0.1\linewidth]{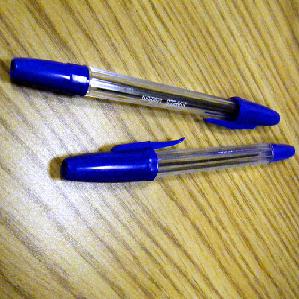}\hfill
    \includegraphics[width=0.1\linewidth]{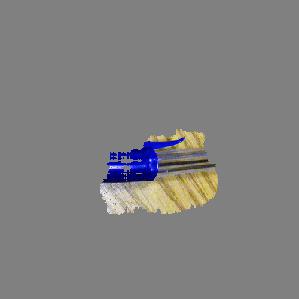}
  }\hspace{0.cm}
  \subfloat[\lab{`bolo tie'}]{
    \includegraphics[width=0.1\linewidth]{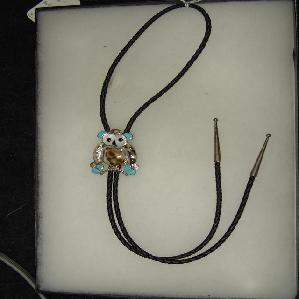}\hfill
    \includegraphics[width=0.1\linewidth]{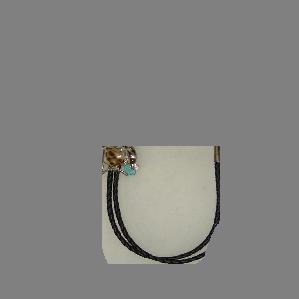}
  }\hspace{0.cm}
  \subfloat[\lab{`projector'}]{
    \includegraphics[width=0.1\linewidth]{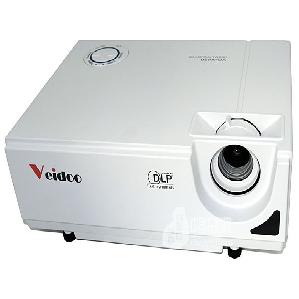}\hfill
    \includegraphics[width=0.1\linewidth]{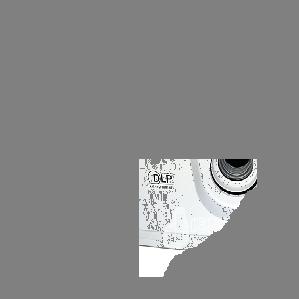}
  }\hspace{0.cm}
  \caption{Input images and explanations from \deepcover for Xception (red labels highlight misclassification or counter-intuitive explanations)} 
  \label{fig:xception}
\end{minipage}\hfill
\begin{minipage}{.4\textwidth}
  \centering
  \begin{tabular}{@{\hspace*{-0.25cm}}c@{\hspace{0.25cm}}c@{\,\,}c@{\,\,}c@{\,\,}c}
    \includegraphics[width=0.16\linewidth]{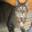}&
    \includegraphics[width=0.16\linewidth]{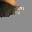}&
    \includegraphics[width=0.16\linewidth]{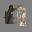}&
    \includegraphics[width=0.16\linewidth]{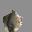}&
    \includegraphics[width=0.16\linewidth]{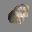}\\
    \includegraphics[width=0.16\linewidth]{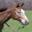}&
    \includegraphics[width=0.16\linewidth]{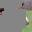}&
    \includegraphics[width=0.16\linewidth]{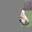}&
    \includegraphics[width=0.16\linewidth]{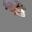}&
    \includegraphics[width=0.16\linewidth]{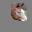}\\
    \includegraphics[width=0.16\linewidth]{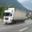}&
    \includegraphics[width=0.16\linewidth]{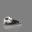}&
    \includegraphics[width=0.16\linewidth]{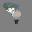}&
    \includegraphics[width=0.16\linewidth]{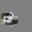}&
    \includegraphics[width=0.16\linewidth]{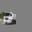}\\
    \includegraphics[width=0.16\linewidth]{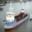}&
    \includegraphics[width=0.16\linewidth]{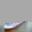}&
    \includegraphics[width=0.16\linewidth]{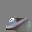}&
    \includegraphics[width=0.16\linewidth]{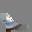}&
    \includegraphics[width=0.16\linewidth]{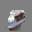}\\
    \includegraphics[width=0.16\linewidth]{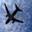}&
    \includegraphics[width=0.16\linewidth]{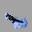}&
    \includegraphics[width=0.16\linewidth]{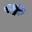}&
    \includegraphics[width=0.16\linewidth]{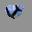}&
    \includegraphics[width=0.16\linewidth]{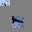}\\
    \scriptsize\textbf{Original}&\scriptsize\textbf{It.\,1}&\scriptsize\textbf{It.\,5}&
    \scriptsize\textbf{It.\,10}&\scriptsize\textbf{It.\,20}
  \end{tabular}
   \caption{Explanations of the DNN at different training stages: the 1st column are the original images
   and the subsequent columns give the explanations for a particular training iteration (CIFAR-10 validation data set)}
  \label{fig:cifar10-e}
\end{minipage}
\end{figure}

Furthermore, we~apply \deepcover after each training iteration to the intermediate DNN. In Fig.~\ref{fig:cifar10-e} we showcase some representative results at different stages of the training.
Overall, as the training procedure progresses, explanations of the DNN's decisions focus more on
the ``meaningful" part of the input image, e.g., those pixels contributing to the interpretation of image (see, for example, the
progress of the training reflected in the explanations of DNN's classification of the first image as a `cat').
This result reflects that the DNN is being trained to learn features of different classes of inputs.
Interestingly, we also observed that the DNN's feature learning is not always monotonic, as
demonstrated in the bottom row of Fig.~\ref{fig:cifar10-e}: after the $10$th iteration, explanations for the DNN's classification of an input image as an `airplane' drift away from the intuitive parts of the input towards pixels that may not fit human interpretation (we repeated the experiments multiple times to minimize the uncertainty because of the randomization in our SFL algorithm). 
The explanations generated by \deepcover may thus be useful for assessing 
the adequacy of the DNN training: they allow us to check, whether the DNN is aligned
with the developer's intent during training. % and thus be used as a stopping
%condition for the training process. %: training is finished when the explanations align with our intuition.
Additionally, the results in Fig.~\ref{fig:cifar10-e} satisfy the 
``sanity'' requirement postulated in~\cite{adebayo2018sanity}: the explanations from \deepcover evolve when the model parameters change during the training.

\subsection{Comparison with the state-of-the-art}

We compare \deepcover with state-of-the-art DNN explanation tools. 
The DNN is VGG16 and we randomly sample 1,000 images from ILSVRC2012
as inputs. We evaluate the effect of highly ranked features by different methods following
an addition/deletion style experiment~\cite{petsiuk2018rise,fong2019understanding}.

An explanation computed by Algorithm~\ref{algo:sbe} is a subset $\mathcal{P}^\mathit{exp}$ of top-ranked pixels 
out of the set $\mathcal{P}$ of all $224$$\times$$224$ pixels that is sufficient for the DNN to classify the image correctly.
We define the size of the explanation as $\frac{|\mathcal{P}^\mathit{exp}|}{|\mathcal{P}|}$. We use the size of an explanation as a proxy for its quality.
%the smaller this size is,
%the more accurately we captured the decision process of the DNN, hence smaller explanations are considered better.

%hana changed the size of figures to 0.5
%\begin{figure}[!htb]
%  \centering
%  \includegraphics[width=.5\columnwidth]{images/ijcai/addition.pdf}
%  \caption{Comparison in the size of generated explanations by different tools on VGG16}
%  \label{fig:restore}
%\end{figure}

\begin{figure}
\centering
\begin{minipage}{.48\textwidth}
  \centering
  \includegraphics[width=.99\columnwidth]{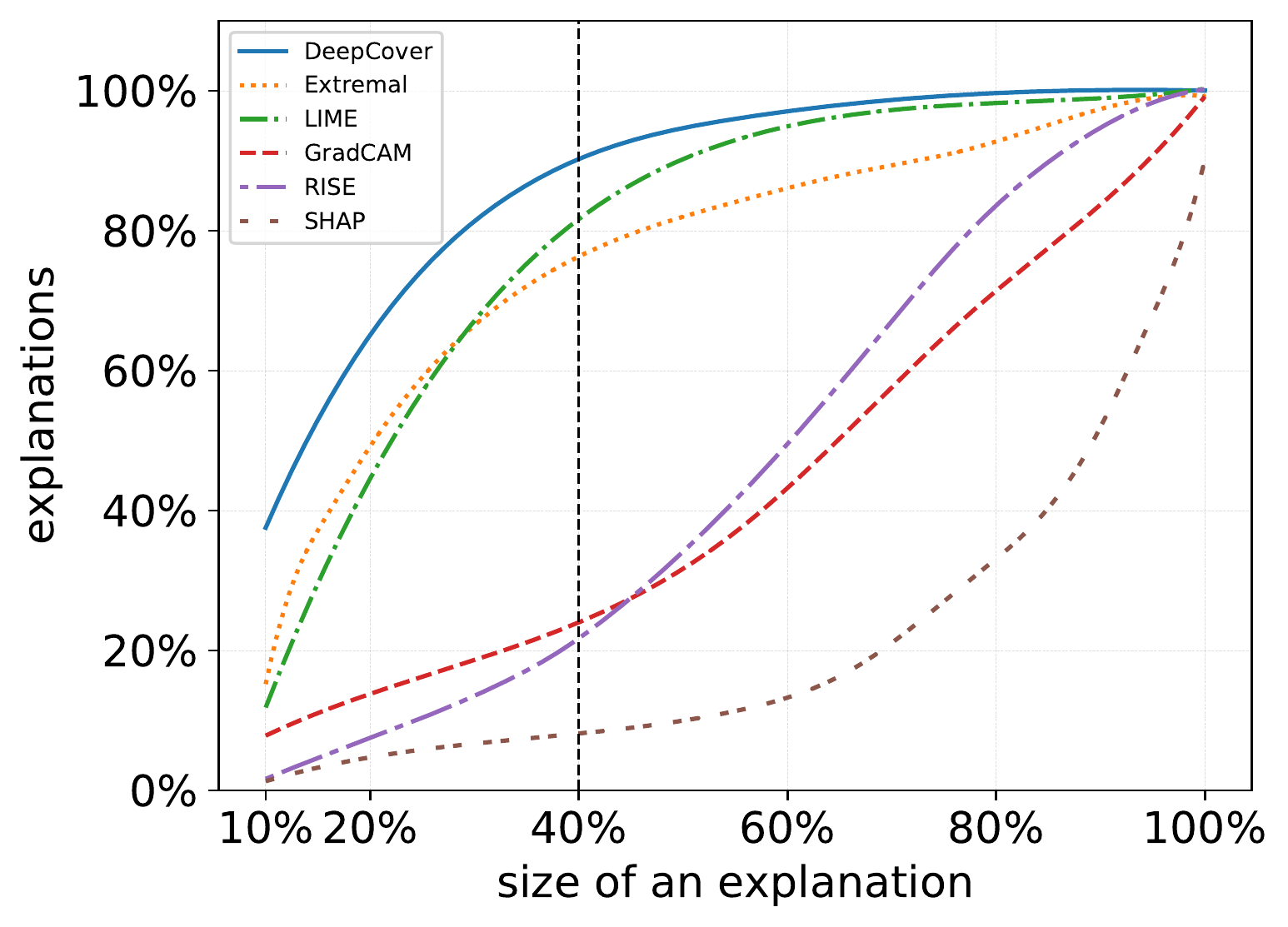}
  \caption{Comparison in the size of generated explanations by different tools}
  \label{fig:restore}
\end{minipage}\hfill
\begin{minipage}{.48\textwidth}
  \centering
  \includegraphics[width=.99\columnwidth]{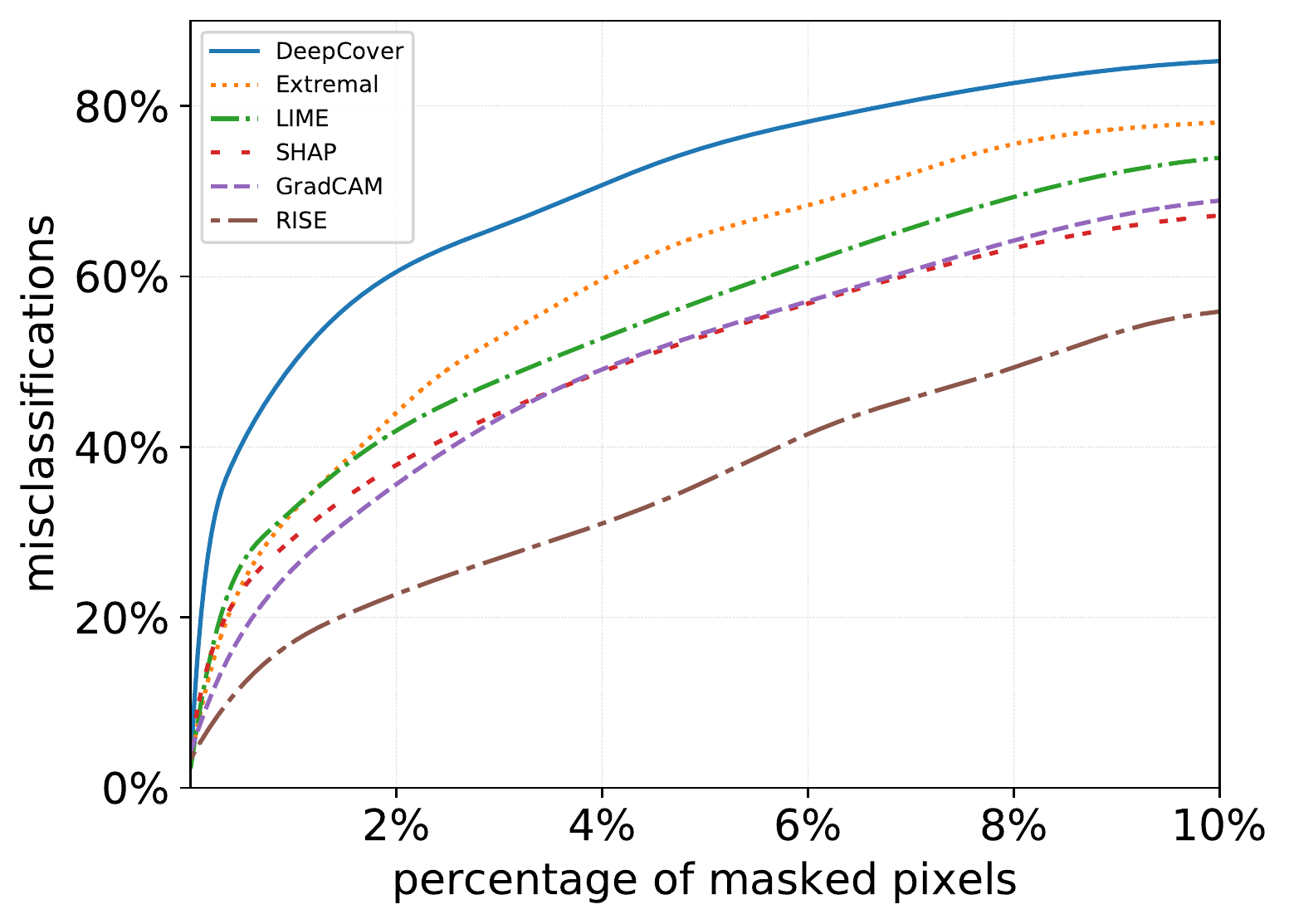}
  \caption{Misclassification vs percentage of masked pixels for different tools}
  \label{fig:attack}
\end{minipage}
\end{figure}

Fig.~\ref{fig:restore} compares \deepcover and its competitors with respect to
the size of the generated explanations. The position on the $x$-axis is the size of the explanation,
and the position on the $y$-axis gives the accumulated percentage of explanations:
that is, all generated explanations with smaller or equal size.

The data in Fig.~\ref{fig:restore} suggests that explanations based on SFL ranking are superior
in terms of their size. For example, nearly $40\%$ of the DNN inputs can be explained
via \deepcover using no more than $10\%$ of the total input pixels, which is two times as good as the
second best explanation method \extremal.

We quantify the degree of redundancy in the generated explanations as follows. We mask pixels following the ranking generated by the different methods until we obtain a different classification. The smaller the number of pixels that have to be masked, the more important the highest-ranked features are. We present the number of pixels changed (normalized over the total number of pixels) in Fig.~\ref{fig:attack}.  Again, \deepcover dominates the others. Using \deepcover's ranking, the classification is changed after masking no more than $2\%$ of the total pixels in $60\%$ of the images. To achieve the same classification outcomes, the second best method \extremal requires changing $4\%$ of the total number of pixels, and that is twice the number of pixels needed by \deepcover.

\begin{figure}
\begin{minipage}{.4\textwidth}
  \centering
  \includegraphics[width=.9\columnwidth]{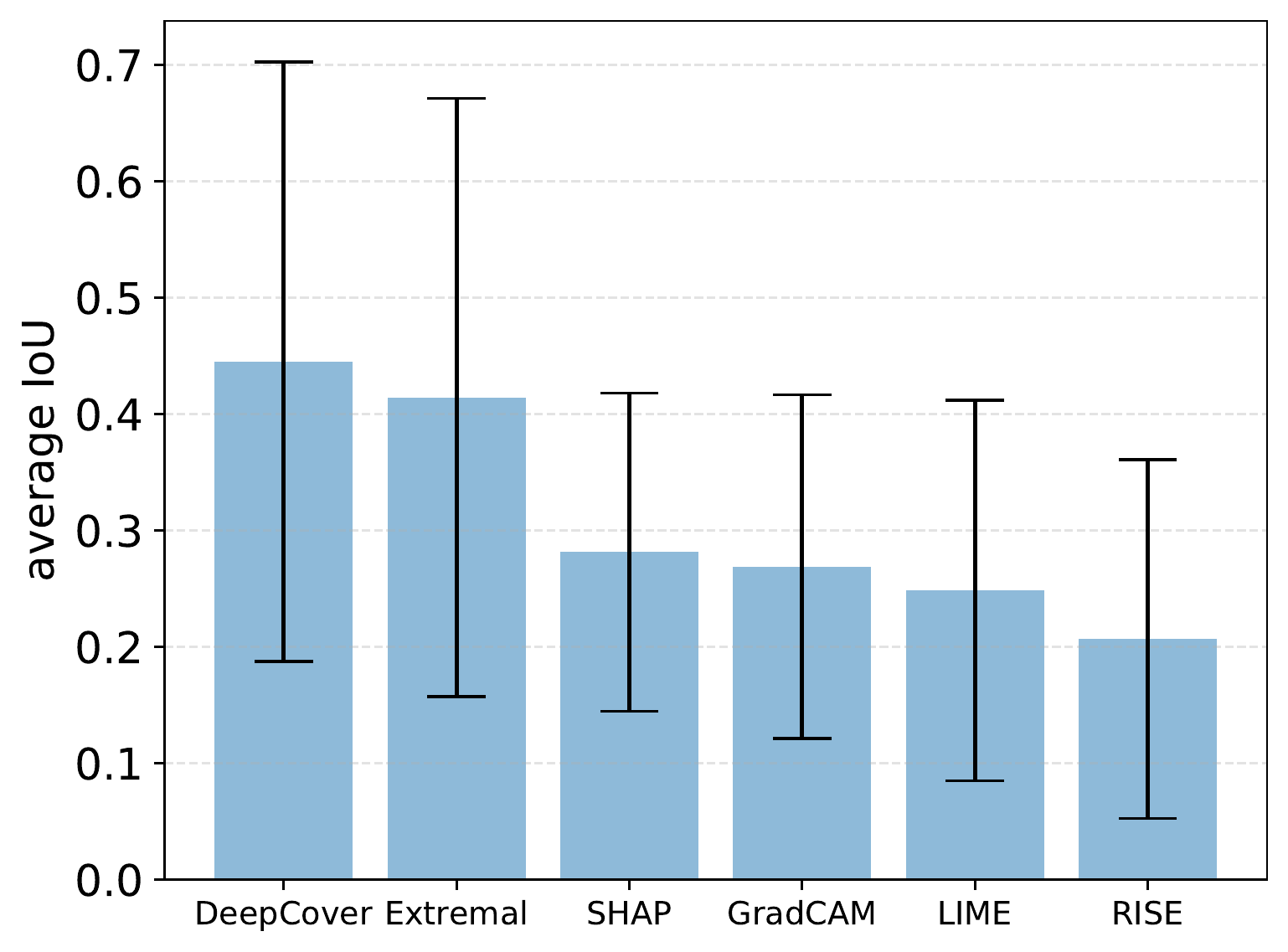}
  \caption{Results for weakly supervised object localisation}
  \label{fig:wsol}
\end{minipage}\hspace{0.05\textwidth}
\begin{minipage}{.5\textwidth}
  \centering
  \begin{tabular}{@{\hspace*{-0.25cm}}c@{\hspace{0.25cm}}c@{\,\,}c@{\,\,\,\,}c@{\,\,}c}
    \includegraphics[width=0.2\linewidth]{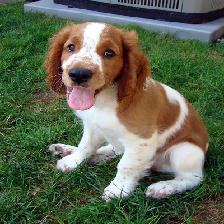}&
    \includegraphics[width=0.2\linewidth]{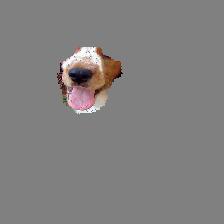}&
    \includegraphics[width=0.2\linewidth]{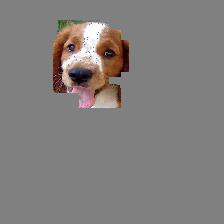}&
    \includegraphics[width=0.2\linewidth]{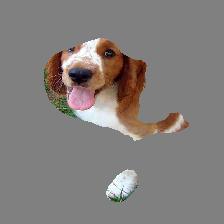}&
    \includegraphics[width=0.2\linewidth]{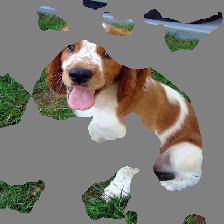}\\
    \tiny{Original}& \tiny{n=2,000} & \tiny{n=200} & \tiny{n=2,000} & \tiny{n=200} \\
    &     \multicolumn{2}{c}{\scriptsize{\deepcover}} & \multicolumn{2}{c}{\scriptsize{\rise}}\\
  \end{tabular}
   \caption{Explanations for the {\lab`Welsh springer spaniel'} by \deepcover and \rise with varying number
   of samples (i.e. $n$)}
  \label{fig:comparison-rise}
\end{minipage}
%\begin{minipage}{.3\textwidth}
%  \centering
%  \includegraphics[width=.9\columnwidth]{images/evaluation/restore-mobilenet-bar.pdf}
%  \caption{The size of the explanation for \deepcover's four measures when its mutant set size $|T(x)|$ varies}
%  \label{fig:change-size}
%\end{minipage}
\end{figure}

\paragraph{Discussion}
We have refrained from using human judges to assess the quality of the explanations, and instead
use size as a proxy measure to quantify the quality of explanation. However, a smaller explanation is not necessarily a better explanation---in fact, ``people have a preference for explanations with some redundancy''~\cite{zemla2017evaluating}.  We therefore complement our evaluation with further experiments. Fig.~\ref{fig:wsol} gives the results of using the explanations for the weakly supervised object localization (WSOL).
We measure the intersection of union (IoU) between the object bounding box and the equivalent number of top-ranked pixels. The IoU is a standard measure of success in object detection and a higher IoU is better. The results confirm again that the top-ranked pixels from \deepcover perform better than those generated by other tools.

%hana changed the size of figure to 0.5
%\begin{figure}[!htb]
%  \centering
%  \includegraphics[width=.5\columnwidth]{images/ijcai/deletion-crop.pdf}
%  \caption{Misclassification vs percentage of masked pixels for different tools on VGG16}
%  \label{fig:attack}
%\end{figure}

\paragraph{Comparison with Rise} 
The \rise tool generates random masks and calculates a ranking of the input pixels using the expected confidence of the classification of the masked images. A rank of a pixel $p$ by \rise depends only on the
confidence of the images in which $p$ is unmasked. 
%\rise ranks pixels by generating random masks using Monte Carlo sampling and then passes masked input
%images to the classifier. 
By contrast, \deepcover uses a binary classification (a mutant image is either classified the same as the
original image or not) and takes into account both the images where $p$ is masked and where it is unmasked.
Figs.~\ref{fig:restore} and~\ref{fig:attack} demonstrate that \deepcover outperforms \rise, producing
smaller and more intuitive explanations.
%In addition to the better performance shown in Fig.~\ref{fig:restore}
%and \ref{fig:attack}, there are at least two further advantages of \deepcover. 
Furthermore, the \deepcover approach
is more general and does not depend on a particular sampling distribution as long as
its mutant test suite is balanced (Sec.~\ref{sec:sfl-explanation}).
%whereas \rise relies on Monte Carlo sampling. 
%We believe that the RISE approach can be integrated into
%the SFL framework. 
Moreover, the \deepcover approach is less sensitive to the size of the mutant test suite (Fig.~\ref{fig:comparison-rise}). 
When the size of the test suite decreased from 2,000 to 200, the size of the generated explanation
only increased by $3\%$ of the total pixels on average. % with respect to the results in Fig. \ref{fig:restore}.
%Fig.~\ref{fig:change-size} further confirms this claim, showing that if the sample size is decreased from $\mbox{2,000}$ to $200$, the performance loss of \deepcover is insignificant.

Next, we present a synthetic benchmark (Sec.~\ref{sec:chimera}) and a security application (Sec.~\ref{sec:trojaning}).

\subsection{Generating ``ground truth'' with a Chimera benchmark}
\label{sec:chimera}
\begin{figure}
\centering
\begin{minipage}{.3\textwidth}
  \centering
  \includegraphics[width=0.35\linewidth]{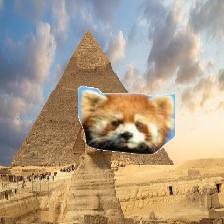}\hspace{0.1cm}
  \includegraphics[width=0.35\linewidth]{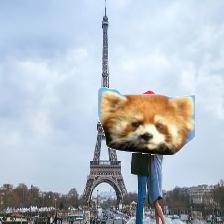}\hspace{0.1cm}
  \includegraphics[width=0.35\linewidth]{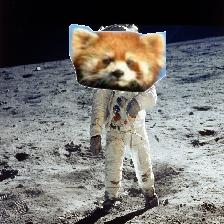}\hspace{0.1cm}
  \includegraphics[width=0.35\linewidth]{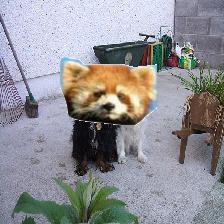}
  \caption{Examples of embedding the red panda}
  \label{fig:chimera}
\end{minipage}\hfill
\begin{minipage}{0.6\textwidth}
  \centering
    %\begin{tabular}{@{}lcccc@{}}
    %\toprule
    %                    & \deepcover            &       \extremal &          \rise  & \gradcam \\ \cline{2-5}
    % IoU$\geq$0.5       & \textbf{76.7\%}       &          70.7\% &         55.6\%  &  0\%  \\
    % IoU$\geq$0.6       & \textbf{54.9\%}       &          21.5\% &         44.4\%  &  0\%  \\
    % IoU$\geq$0.7       & \hphantom{0}9.8\%     &\hphantom{0}2.2\%& \textbf{28.2}\% &  0\%  \\
    %\bottomrule
    %\end{tabular}
    \begin{tabular}{@{}l@{\qquad}r@{\qquad}r@{\qquad}r@{}}
    \toprule
                        &  IoU$\geq$0.5     &   IoU$\geq$0.6        &   IoU$\geq$0.7        \\ \cline{2-4}
     \deepcover         & \textbf{76.7\%}   &   \textbf{54.9\%}     &   \hphantom{0}9.8\%   \\
     \extremal          &  70.7\%           &   21.5\%              &   \hphantom{0}2.2\%   \\
     \rise              &  55.8\%           &   42.9\%              &   \textbf{25.7}\%     \\
     \gradcam           &  \hphantom{0}\hphantom{0}0\%           &   \hphantom{0}\hphantom{0}0\%              &   \hphantom{0}\hphantom{0}0\%     \\
     
    \bottomrule
    \end{tabular}
    \captionof{table}{IoUs between the embedded red panda and the highest ranked pixels
    for four different tools}
    \label{tab:chimera-iou}
\end{minipage}
\end{figure}

The biggest challenge in evaluating explanations for DNNs (and even for human decision making)
is the lack of the \emph{ground truth}. Human evaluations of the explanations
remain the most widely accepted measure, but are often subjective. 
In the experiment we describe below, we synthesize a \emph{Chimera benchmark}\footnote{The benchmark images are publicly available at \url{http://www.roaming-panda.com/}.} by randomly superimposing
a ``red panda'' explanation (a part of the image of the red panda) onto a set of randomly chosen images. The benchmark consists of
$1,000$ composed (aka ``Chimera'') images that retain the ``red panda'' label when using both the MobileNet and the VGG16 classifiers. Fig.~\ref{fig:chimera} gives several examples of the Chimera images. The rationale is that if such an image is indeed classified as ``red panda'' by the DNN, then the explanation of this classification must be contained among the pixels we have superimposed onto the original image.

For each image from the Chimera benchmark, we rank its pixels using \deepcover and other tools.
We then check whether any of their top-$\pi$ highest ranked pixels are part of the ``red panda''.
In Table~\ref{tab:chimera-iou}, we measure the IoU (intersection of union) between the ground truth explanation
and the top-$\pi$ highest ranked pixels, where $\pi$ ranges from $1\%$ to $100\%$.
Assuming that an IoU $\geq 0.5$ is a successful detection, \deepcover successfully detects the
ground truth planted in the image in $76.7\%$ of the total cases and it is $6\%$
better than the second best \extremal. The benefit provided by \deepcover is even more substantial when
requiring 0.6 IoU. Overall, the results in Table~\ref{tab:chimera-iou} are
consistent with the addition/deletion experiment (Figs.~\ref{fig:restore} and~\ref{fig:attack}) and the WSOL
experiment (Fig.~\ref{fig:wsol}), with \deepcover topping the list. Interestingly, when \rise succeeds to find the explanation, it seems to localize it better (with IoU $\geq 0.7$). \gradcam fails to detect the embedded red panda in all cases. These observations support the hypothesis that a benchmark like Chimera is a good approximation for ground truth, and helps us to compare algorithmic alternatives.

\subsection{Trojaning attacks}
\label{sec:trojaning}

\begin{figure}[!htb]
  \centering
  \subfloat[(a)]{
    \includegraphics[width=0.15\linewidth]{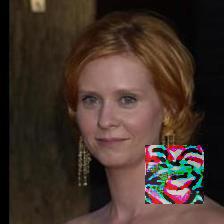}\hfill
    \includegraphics[width=0.15\linewidth]{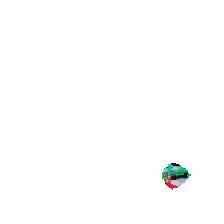}
  }\hspace{1.cm}
  \subfloat[(b)]{
    \includegraphics[width=0.15\linewidth]{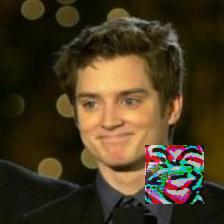}\hfill
    \includegraphics[width=0.15\linewidth]{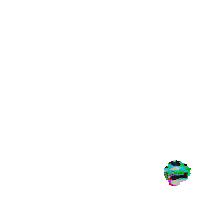}
  }\hspace{0.cm}
  \caption{Applying \deepcover to Trojaning attacks on VGG Face. The Trojan trigger is the square shape in the lower right corner of the image; 
  the \deepcover explanation for the Trojan behaviour is on the right.}
  \label{fig:trojaning}
\end{figure}

The authors of~\cite{trojaning} say that a DNN is  ``trojaned'' if it behaves correctly on ordinary input images but exhibits malicious behavior when a ``Trojan trigger'' is part of the input. Thus, we can treat this trigger as a ground truth explanation for the Trojaned behavior of the DNN. We have applied \deepcover to identify the embedded trigger
in the input image for the Trojaned VGG Face~\cite{trojaning}. The result is illustrated in Fig.~\ref{fig:trojaning}.
This use case suggests that there is scope for the application of  \deepcover in DNN security.

When applying \deepcover to the Trojaned data set in~\cite{trojaning}, 
the top $8\%$ highest ranked pixels have an average IoU value of~0.6 with the Trojan trigger. 
According to \deepcover, the Trojaning output for each input is caused by a small part of its embedded
trigger. This black-box discovery by \deepcover is consistent with and further optimizes the theory of DNN Trojaning~\cite{trojaning}.
Finally, as many as $80\%$ of the (ground truth) Trojan triggers are successively localized (with IoU $\geq 0.5$) by only $\pi=8\%$ of the pixels top-ranked by \deepcover. \deepcover is thus very effective.

\subsection{Threats to Validity}
In this part, we highlight several threats to the validity of our evaluation.
\paragraph*{Lack of ground truth}
We have no ground truth for evaluating the generated explanations for Xception on ImageNet images, hence we use the size
of an explanation as a proxy. We have the ground truth for the Chimera images of red panda (Fig.~\ref{fig:chimera})
and for the Trojaning attacks (Fig.~\ref{fig:trojaning}), and the results support our claims of the high quality of \deepcover explanations.

%, the size of the explanation and the effort required for generating adversarial examples.

%\paragraph*{Selection of the dataset}
%In this paper, we focus on the image recognition problem for high-resolution color images and collect most of the experimental results using the ImageNet data set. Small benchmarks and problems may have their own features that differ from what we report in this paper. It is known that, in traditional software, the performance of different 
%spectrum-based measures can vary dramatically given the benchmark used. %SHAP has been applied to DNNs with non-image input.

\paragraph*{Selection of SFL measures}
We have only evaluated four SFL measures (Ochiai, Zoltar, Tarantula and Wong-II). There are hundreds more such measures, which may reveal new observations.

\paragraph*{Selection of parameters when generating test inputs}
When generating the test suite $T(x)$, we empirically configure the parameters in the test generation algorithm. The choice of parameters affects the results of the evaluation and they may be overfitted.

\section{Conclusions}
\label{sec:conclusions}

This paper advocates the application of statistical fault localization (SFL) for the generation of explanations of the output of neural networks.
Our definition of explanations is inspired by actual causality, and we demonstrate that we can efficiently compute a good approximation
of a precise explanation using a lightweight ranking of features of the input image based on SFL measures.
The algorithm is implemented in the tool \deepcover. Extensive experimental results demonstrate that 
\deepcover consistently outperforms other explanation tools and that its explanations
are accurate when compared to ground truth (that is, the explanations of the images
have a large overlap with the explanation planted in the image).

\clearpage
\bibliographystyle{splncs04}
\bibliography{all}

%\appendix
%\input{appendix}

\end{document}